\documentclass[10pt,twocolumn,letterpaper]{article}

\usepackage{cvpr}              % To produce the CAMERA-READY version
\definecolor{cvprblue}{rgb}{0.21,0.49,0.74}
\usepackage[pagebackref,breaklinks,colorlinks,allcolors=cvprblue]{hyperref}
\usepackage[table]{xcolor}
\usepackage{multirow}
\usepackage{amsmath}
\def\confName{CVPR}
\def\confYear{2026}

\title{\model: Efficient 3DGS Rendering for Large-Scale Scenes with \\ Inter-frame Caching and Tile Scheduling}
\author{
Tingjia Zhang$^{1*}$, 
Bo Chen$^{2}$, 
Shengzhong Liu$^{1\dagger}$, 
Fan Wu$^{1}$, 
Guihai Chen$^{1}$\\
$^{1}$Shanghai Jiao Tong University,$^{2}$University of Illinois Urbana-Champaign\\
{\small $^{*}$First Author \hspace{10pt} $^{\dagger}$Corresponding Author}
}
\allowdisplaybreaks[4]

\def \eg {\emph{e.g.}, }
\def \ie {\emph{i.e.}, }

\usepackage[linesnumbered, ruled]{algorithm2e}
\def\BibTeX{{\rm B\kern-.05em{\sc i\kern-.025em b}\kern-.08em
    T\kern-.1667em\lower.7ex\hbox{E}\kern-.125emX}}

\SetCommentSty{mycommfont} 

\newcommand{\model}{{CaT-GS}\xspace}

\begin{document}
% Scheduling and Content-Aware
%\title{Title}

%\input{others/authors}

\maketitle

\begin{abstract}
Recent breakthroughs in 3D Gaussian Splatting (3DGS) have advanced neural rendering with high fidelity and efficiency. However, its performance degrades severely in large-scale scenes due to the increasing computational workload of tile-based rasterization. 
Existing acceleration approaches either require costly scene re-training or focus only on the rasterization stage of the pipeline, overlooking general pipeline redundancy in real-time rendering.
Through a comprehensive analysis, we identify three primary sources of redundancy and low GPU utilization: 1) redundant inter-frame pre-processing, 2) viewpoint redundancy, and 3) imbalanced tile load distribution. 
To address these issues, we propose \model, a novel and efficient 3DGS rendering pipeline. \model introduces a speculative multi-frame preprocessing method to eliminate redundant computations across consecutive frames, and an inter-frame caching mechanism to eliminate viewpoint redundant rendering stages.
Furthermore, it redistributes the rasterization tasks with a dedicated CUDA kernel to mitigate tile load imbalance and boost GPU utilization. 
Extensive experiments on large-scale scenes demonstrate that \model achieves a speedup of up to 10× over the original 3DGS and up to 70\% over previous state-of-the-art (SOTA) methods, establishing a new benchmark for high-fidelity, real-time rendering of large-scale scenes.
\end{abstract}

%\bo{narrow aspects are vague}

%\bo{sequential here is a bit confusing without definition. Maybe we just say we identify the redundancy and GPU underutilization in existing works can detail then next}.
%\shengzhong{Be more explicit in pointing out existing limitations. What kind of deficiencies are we referring to?}%\bo{we can highlight that we experiment over a broad specturm of scene sizes}
%\bo{it is a bit strange if we mentioned three sources of problems but proposes only two solutions. Can we separate them into three or we say there are two sources of problems?}

%\bo{maybe we should stress that we focus on a realistic viewing trace instead of testing on a sequence of non-continous images, which raises the following problems?}

% \begin{IEEEkeywords}
% Real-Time Scheduling, Object Detection, Temporal Correlations, Cyber-Physical Systems
% \end{IEEEkeywords}

\section{Introduction}\label{sec:intro}
Driven by the recent breakthrough of 3D Gaussian Splatting (3DGS)~\cite{kerbl20233d}, 3D reconstruction and neural rendering have exhibited significant potential across various interactive applications, including autonomous driving ~\cite{liu2025omnireason,hess2025splatad,zhou2024drivinggaussian}, immersive streaming~\cite{shi20253d,wu2025advancing,chen20253dgv}, and virtual tours ~\cite{sun2025streaming,kwon2025realistic}.
%\bo{virtual world is not very clear. Is it metaverse or just gaming?}
%\bo{autonomous driving is good, but the others vague. Try something like healthcare, education,...}
While 3D Gaussian Splatting (3DGS) represents a significant advance over Neural Radiance Fields (NeRF) ~\cite{mildenhall2021nerf} in both reconstruction quality and rendering speed, achieving high frame rates remains challenging for certain real-world applications, such as virtual world and large-scale city reconstruction. These scenarios typically involve larger and more complex scenes compared to standard datasets, which substantially increase computational overhead and rendering latency. Therefore, it is necessary to investigate methods for accelerating 3DGS rendering in large-scale scenes.%\bo{no need to mention redundancy here, which is part of our solution. "Optimizing" is a vague term. How about ...methods }
% In real-world application, 
 
%3DGS uses Gaussian representations for scenes and employs tile-based rasterization%\bo{while I understand this sentence, I'm not sure if it makes sense to researchers not working with 3DGS. For example, "Gaussian representations" can be confusing. Maybe look at other papers and see how they describe 3DGS.}
%, which incurs a significant computational workload as the number of Gaussians and their interactions with tiles increase substantially with scene complexity.\bo{maybe we can highlight explicitly that realistic rendering involves complex/large scenes to motivate our approach. Then, we can talk about the redundancy and underutilization as our findings to solve this complexity problem.}
%3DGS uses Gaussian representations for scenes and employs tile-based rasterization, when the scene complexity increases, the workload in all stages of 3DGS stages of pre-processing, sorting and rasterizing increase%which incurs a significant computational workload as the number of Gaussians and their interactions with tiles increase substantially with scene complexity.
%\shengzhong{We put large-scale scenes into the title of the paper, then we need to carefully analyze why it is different and what unique challenges to 3DGS rendering that your solution is proposed to solve.}

To enhance 3DGS performance, several existing works~\cite{liu2024compgs,mallick2024taming} aim to reduce the computational overhead from excessive Gaussians through model compression or pruning. These methods primarily focus on improving the Gaussian model's representation efficiency to reduce storage and Gaussian count, rather than directly optimizing its rendering pipeline. Moreover, these methods typically require scene re-training to generate efficient representations; such requirement for re-training impairs their practicality in real-world deployment.
%\bo{do they need to re-train for every scene and every hardware? If so, we can stress that.}.
% 
Another line of research aims to minimize computational redundancy within the rendering pipeline by analyzing the contribution of each Gaussian to specific tiles during rasterization to reduce the unnecessary computations.
Prior works~\cite{wang2024adr,feng2025flashgs,lu2024scaffold} primarily optimize the inefficiency of redundant Gaussian in rasterization by designing fine-grained Gaussian-tile intersection and overlap determination.  %\bo{"the sufficiency of axis-aligned" is hard to understand}

However, these approaches overlook critical aspects of actual deployment pipelines in applications. First, a significant limitation is their disregard for \textbf{viewpoint coherence}. The reported high-speed rendering results are mostly measured in controlled settings using fixed camera pose sets from predefined datasets, the same poses as used in capturing the ground-truth image of the dataset.
%\bo{are you sure about this? What if the training and testing poses have no overlapping?}
This setup is only suitable for assessing reconstruction quality of the rendering algorithm, but lacks fidelity for benchmarking rendering efficiency under continuous and arbitrary camera trajectories in interactive applications where significant redundancies arise in the pre-processing stage (\eg frustum culling and sorting).
Secondly, previous methods fail to fully address the issue of \textbf{low GPU utilization} caused by load imbalance. 
While Flash-GS~\cite{feng2025flashgs} improves compute-memory balance by overlapping Gaussian loading and computing, it fails to address the low GPU utilization issue caused by tile-level load imbalance. 
Alternatively, ADR-GS~\cite{wang2024adr} attempts to alleviate load imbalance by retraining the model with an additional constraint loss on tile-aware Gaussian distribution; however, this approach leads to a degradation in reconstruction quality.
We conduct a thorough analysis of the rasterization step and identify multiple sources of redundancy in standard rendering pipelines. 
\begin{enumerate}[leftmargin=10pt]
\item \textbf{Inter-Frame Redundancy:} In a typical 3DGS deployment, scenes are rendered as sequential, continuous frames for interactive applications, as the viewing pose changes gradually. When the frame rate is high, consecutive frames contain very similar rasterization information derived from pre-processing. These nearly identical pre-processing computations across frames lead to significant inter-frame redundancy.
%\bo{why sequential and continuous? maybe the fundamental reason is that the .}. 
 %To this end,   

%These nearly identical pre-processing computations across a batch of frames lead to significant inter-frame redundancy.  
% 
%\item \textbf{Rasterization View-Point Redundancy:} Although the initial rasterization already constrains Gaussians to those within the camera's frustum, a further subset can be culled. These are Gaussians whose cumulative radiance exceeds the maximum depth threshold, making them effectively occluded and useless for the final image.  
% 
\item \textbf{Uneven Tile Distributions:} 3DGS achieves high rasterization efficiency through its tile-based approach, which partitions the image and renders multiple tiles concurrently to leverage the parallel architecture of modern GPUs. However, in large and complex scenes, the distribution of Gaussian numbers across tiles can be highly biased.
This leads to a severe load imbalance among GPU units during rasterization, resulting in overall GPU underutilization.
\end{enumerate}

We therefore propose \model, the first 3DGS rendering pipeline that optimizes the interactive rendering process for large-scale scenes.
To this end, \model introduces novel speculative multi-frame pre-processing and an inter-frame caching mechanism, which collectively reshape the rendering pipeline to mitigate inter-frame redundancy. Furthermore, \model incorporates a dynamic load redistribution strategy and an optimized rendering kernel to comprehensively solve the tile imbalance problem. Through evaluations on models from a large-scale UAV city reconstruction dataset as well as standard benchmarks, \model achieves up to $10\times$ speedup over the vanilla 3DGS pipeline and up to $70\%$ speedup against existing SOTA methods.

Our main contributions in this paper include: 
%\shengzhong{Missing a paragraph to summarize the evaluation setups and results?}

\begin{itemize}
\item We design a speculative rasterization pre-processing method that enables batched pre-processing for sequential frames to avoid full recomputations.
\item We develop an inter-frame caching mechanism to reuse viewpoint-invariant features in 3DGS rendering, thereby bypassing redundant stages from the original pipeline.
\item We alleviate tile load imbalance in 3DGS rasterization by proposing a task-splitting strategy and a dedicated CUDA rendering kernel to improve the GPU utilization during rasterization.
\item \model achieves up to $10\times$ speed up compared to vanilla 3DGS pipeline and up to $70\%$ speed up against existing SOTA baselines.
\end{itemize} 

%\bo{how about dynamic load redistribution?} 

%We therefore propose \model, an efficient 3DGS rendering pipeline that eliminates inter-frame redundancy through a novel caching mechanism and optimizes GPU utilization by refactoring computational tasks to align with modern GPU architectures. 
%\model introduces key innovations to address deficiencies in the standard 3DGS pipeline. 
% 
%First, it employs a speculative pre-processing stage that computes rasterization information for a batch of frames upfront to eliminate redundant per-frame computation. 
% 
%Second, an inter-frame caching mechanism is designed to bypass repetitive steps like sorting and frustum culling across frames, sharing viewpoint-invariant data to further reduce overhead. 
% 
%Finally, \model introduces a task re-factorizing strategy, supported by a modified render kernel, which effectively tackles tile-level load imbalance, thereby improving GPU utilization. 
%\shengzhong{I feel this paragraph can not sufficiently address why the proposed approach is novel and effective. We should spend more effort in highlighting the unique perspective of the idea we are proposing. Summarizing its main differences and advantages over the baselines. Make the analysis above shorter and expand the intuition and insight of our approach.}

\section{Preliminaries and Background}\label{sec:motivate}
\subsection{3DGS Rendering}

\begin{figure}
    \centering
    \includegraphics[width=1.09\linewidth]{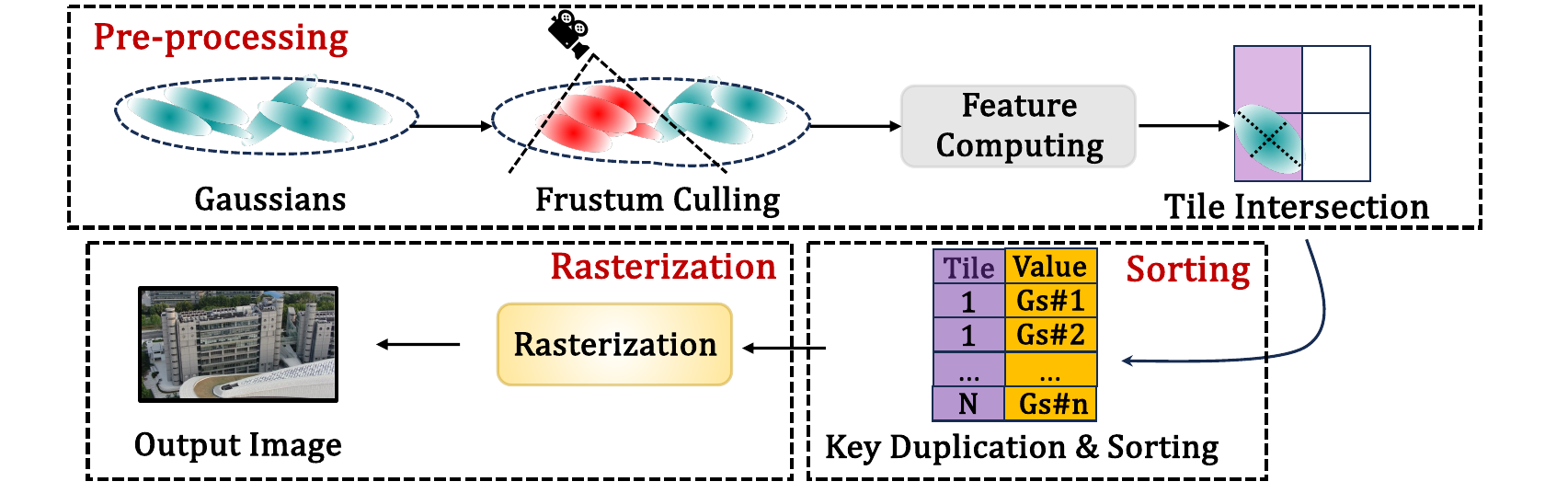}
    \caption{General 3DGS Rendering Pipeline, composed by pre-processing, sorting and rasterization stage.}%\bo{feature computing is not mentioned in the paragraphs and preprocessing is not highlighted in the figure. They should be consistent.}}
    \label{fig:3dgspipe}
\end{figure}

The real-time rendering pipeline of 3D Gaussian Splatting (3DGS) can be decomposed into three main stages: 1) preprocessing, 2) sorting and 3) rasterization. This pipeline is meticulously designed to leverage parallel computing architectures, particularly modern GPUs.
%\shengzhong{preprocessing or pre-processing? The term should be unified.}

\noindent\textbf{1) Pre-processing:} In the pre-processing stage, 3D Gaussians are prepared for subsequent image-plane projection and rasterization. First, 3DGS applies \textbf{frustum culling} to discard the Gaussians outside the viewing volume and reduce computational load. Then the shape and color feature is estimated with 3D covariance matrix and SH functions.
The 2D covariance matrix $\boldsymbol{\Sigma}'$ is computed from 3D covariance $\Sigma$ 
, which formulates the ellipse shape on the splatted plate.
Since 3DGS renders images in a tile-level parallelism, the screen is divided into fixed-size tiles (\eg $16 \times 16$ pixels). 
Each Gaussian is assigned to all tiles whose projected 2D bounding boxes intersect, which creates an initial list of Gaussians per tile.

\noindent\textbf{2) Sorting:} In the second stage, the associated Gaussians in each tile are sorted by their depth (typically the z-coordinate in the camera space). This ensures correct back-to-front blending during rasterization.

\noindent\textbf{3) Rasterization:} The final rasterization stage generates the output image. The render kernel is parallelized across pixels%\bo{pixel or tile?} 
to utilize the GPU capacity. For each pixel $p$, all Gaussians within its corresponding tile are loaded and processed in from near to far depth order. 
%\shengzhong{Increasing or decreasing with the depth?}
Each Gaussian computes an $\alpha$ value $\alpha_{i}(p)=\sigma_{i}g_{i}(p)$ to measure its contribution, where $g_{i}$ is the contribution of the projected 2D Gaussian at pixel $p$:
\begin{equation}
g_{i}=e^{q},q=-\frac{1}{2}(p-\mu_{i_{2D}})\bf{\Sigma}_{i_{2D}}^{-1}(p-\mu_{i_{2D}})^{T}.
\end{equation}

\noindent If $\alpha_{i}>\frac{1}{255}$, then the Gaussian is included in the alpha compositing of the pixel color $C$. The overall color of a pixel is obtained through alpha blending by color $c_{i}$ and weight $\alpha_i$:
\begin{equation}
\label{eq:3}
C(p)=\sum_{i\in\mathcal{N}}c_{i}\alpha_{i}(p)\prod_{j=1}^{i-1}(1-\alpha_{j}(p)).
\end{equation}
%\shengzhong{What is $c_i$? Be careful that all used symbols should be clearly defined.}

\begin{figure}
    \centering
    \includegraphics[width=\linewidth]{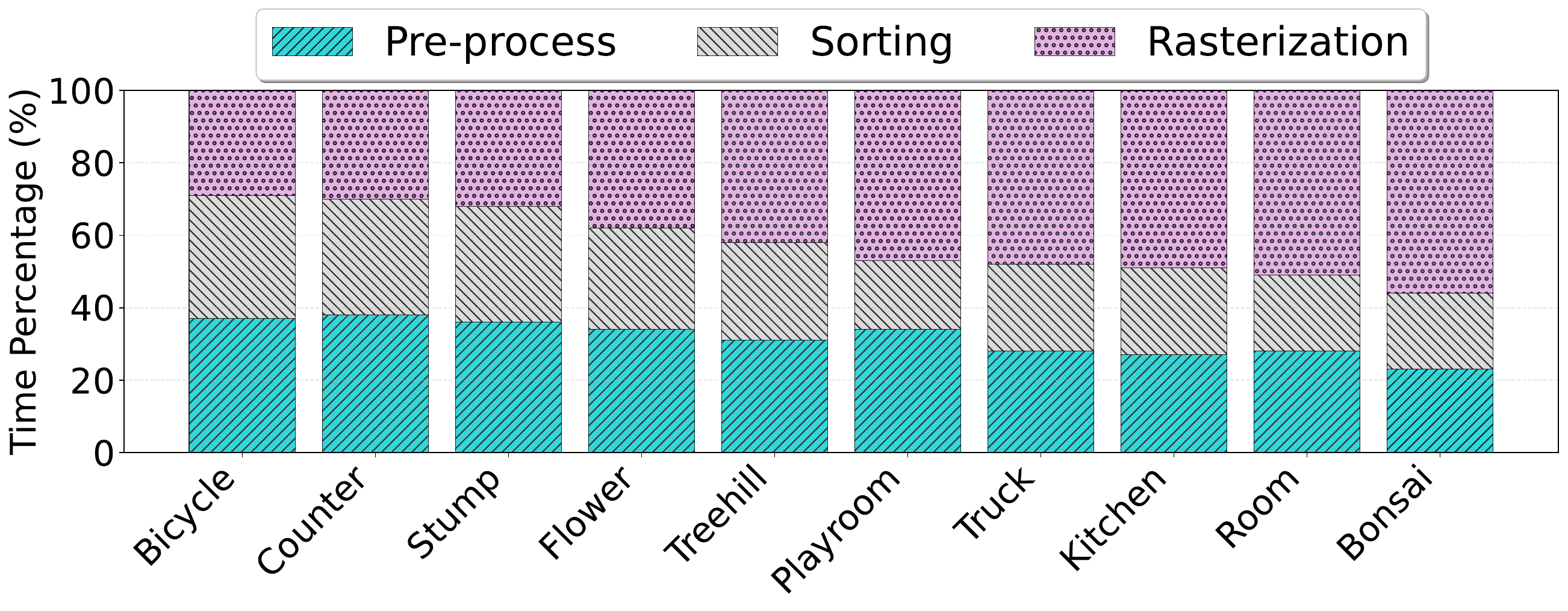}
    \caption{The ratio of rendering time spent on pre-processing, sorting, and rasterization in 3D Gaussian Splatting (3DGS). Note that this ratio depends on both the model and the viewpoint.}
    \label{fig:3dgsratio}
\end{figure}

\subsection{Accelerate 3DGS Rendering}
Previous research has primarily focused on exploiting geometric redundancy to reduce computational overhead. Adr-GS~\cite{wang2024adr} addresses the insufficiency of Gaussian tile intersection determination in vanilla 3DGS by designing a method with tighter axis-aligned bounding boxes (AABBs). This enables more precise intersection judgment and yields a shorter redundant rasterization list. Following this, Flash-GS~\cite{feng2025flashgs} introduced an adaptive intersection judgment mechanism to further compress redundant tasks via opacity pruning. Collectively, these prior works have made a great step in decreasing intra-frame redundancy of 3DGS rendering.%\bo{This sentence is too negative for a paper. Just saying they are intra-frame methods should be enough.}

However, while existing methods extensively explore intra-frame redundancy, few have considered inter-frame redundancy in consecutive rendering.  The limitation is that these methods overlook this inter-frame redundancy, which represents a significant opportunity to reduce overhead in the sorting and pre-processing stages. Yet, as we profile several models from MipNeRF360~\cite{barron2022mip} at random example viewpoint in Figure~\ref{fig:3dgsratio}, the sorting and pre-processing stage also accounts for a significant portion of the rendering time. %\bo{Need a bit details on the profiling like what device and some citation to datasets.}

Alternatively, at the inter-frame level, we identify redundancy in the pre-processing and sorting stages. When the camera moves only slightly, the outcomes of these stages often remain largely unchanged. For instance, frustum culling, which identifies the visible part of the Gaussian model, typically produces similar results across consecutive frames when the rendering frame rate is high. Similarly, the depth-based sorting of Gaussians remains consistent when the camera viewpoint does not change drastically. This presents an opportunity to avoid repetitive computations by processing rendering frames in batches.
%However, previous methods mainly cut down computations and accelerate the rasterization stage, while 
%We begin by profiling the rendering time distribution across three general stages of Flash-GS when rendering different models, as shown in Figure~\ref{fig:3dgsratio}. It can be observed that the time contribution of these three stages varies significantly across scenes and viewpoints. 

%However, while existing methods extensively explore intra-frame redundancy, few have considered inter-frame redundancy in consecutive rendering. Although previous state-of-the-art (SOTA) works claim rendering speeds of over 120 fps or higher, the conventional rendering pipeline, as illustrated in Figure~\ref{fig:3dgspipe}, still exhibits redundancy at the inter-frame level.
%\shengzhong{It seems the results are not related to ``inter-frame redundancy'' but exhibit the heterogeneity among scenes. Why do we want to put it here?}

\begin{figure}
    \centering
    \includegraphics[width=1.0\linewidth]{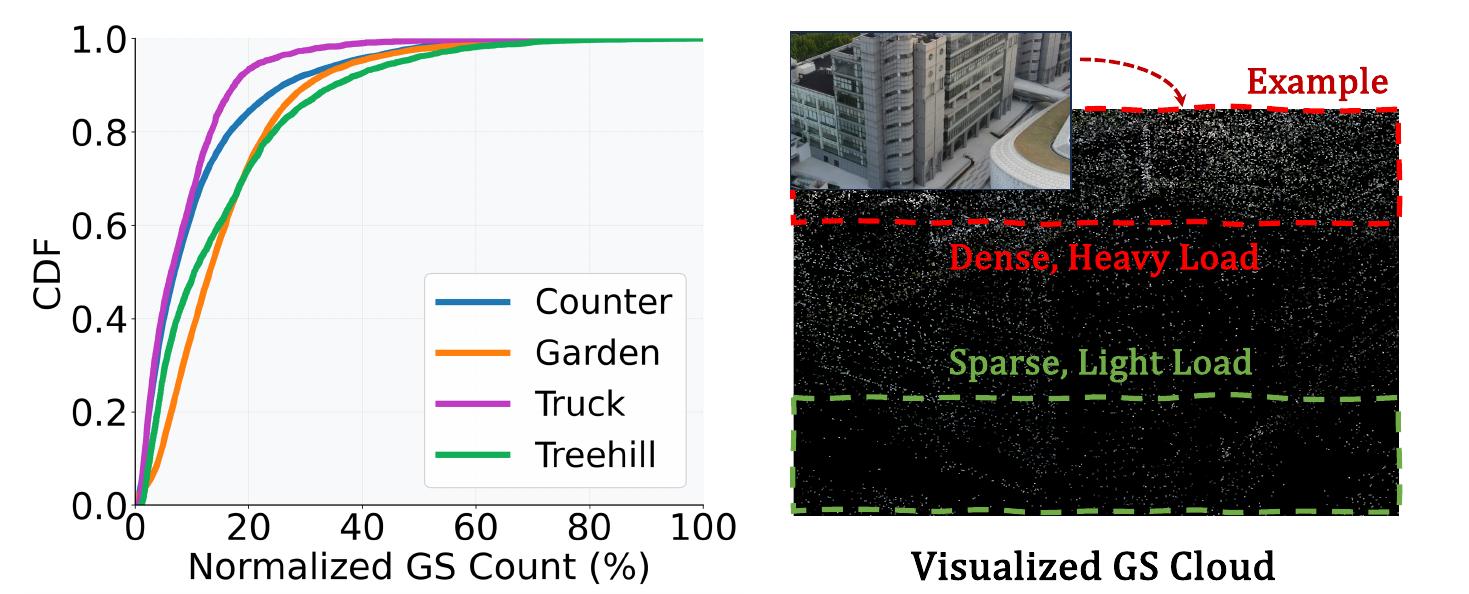}
    \caption{Unbalanced tile-load distribution across scenes. The left figure shows CDF on GS counts per-tile, right figure provides an example of imbalanced load.} %\shengzhong{In the left figure, the x-axis should be unified among different scenes. Otherwise, the readers can not visually capture the heterogeneity of GS count distribution, but misinterpret it as similar distributions.}}
    \label{fig:dist}
\end{figure}

Second, we note remaining inefficiencies in the rasterization stage due to tile load imbalance. This issue arises from the non-uniform distribution of Gaussians among regions. Regions with high-frequency details tend to incur significantly higher rendering loads than others, as depicted in Figure~\ref{fig:dist}. $10\%$ tiles may contain over $50\%$ compute load in some cases.%\bo{List concrete numbers like 20\% tiles contains more than 80\% Gaussians.} 
This load imbalance is evident across various models, as shown by the CDF in the figure. Crucially, the heavy load tiles may cause delay in the parallel rendering. %bo{The key problem is the slowest one will delay all others.}
%\shengzhong{Why are they conflicted? What assumptions are made in the GPU kernel implementations?}

\section{\model Method}\label{sec:framework}

\subsection{Overview}
\begin{figure*}
    \centering
    \includegraphics[width=1.02\linewidth]{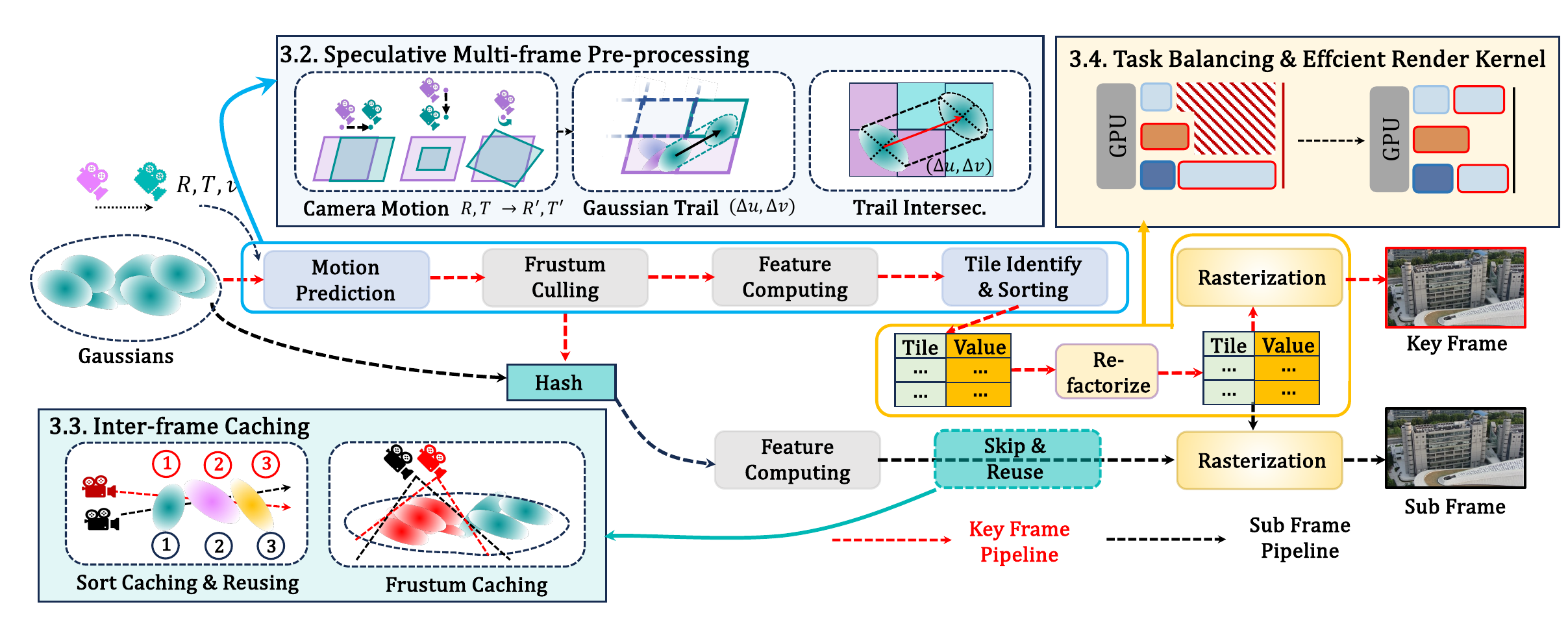}
    \caption{Overview of \model.}
    \label{fig:overview}
\end{figure*}

\model designs a novel pipeline for streamed rendering of 3D Gaussian Splatting (3DGS) in interactive applications, which eliminates inter-frame redundancy and improves GPU utilization. Unlike prior works, \model processes sequential frames in groups. The first frame in a group is designated as the \emph{key frame}, and the subsequent ones as \emph{sub-frames}, with distinct rendering pipelines applied to each type of frame.

As shown by the red line in Figure~\ref{fig:overview}, the key frame pipeline begins with a preprocessing stage that generates rasterization information reusable across the entire frame group. This is achieved through our proposed viewpoint-speculative preprocessing (Section~\ref{sec:3.1}), which performs motion prediction based on viewpoint changes and applies a traced Gaussian intersection judgment to ensure the resulting Gaussian list covers all Gaussians needed throughout the frame group.
In contrast, the sub-frame rendering pipeline skips several stages via an inter-frame caching strategy (Section~\ref{sec:3.2}). When viewpoint movement is small, sub-frames avoid repeated sorting and frustum culling by reusing information from the key frame,  as they remain largely consistent under small viewpoint changes.

To further address the deficiency caused by tile load imbalance and enhance the GPU utilization during rasterization, we introduce a task-balancing method along with a high-performance rendering kernel (Section~\ref{sec:3.4}).

\subsection{Speculative Multi-Frame Pre-Processing}
\label{sec:3.1}
%The rendering pipeline of 3D Gaussian Splatting (3DGS) is initiated by a view-dependent pre-processing stage, which transforms the world-space Gaussian primitives into a renderable form for subsequent tile-based rasterization. First, it performs frustum culling on the Gaussian model to filter the set of 3D Gaussians against the current camera's view frustum. Each Gaussian is projected into 2D screen-space to determine their visibility from the current viewpoint. Second, for the visible Gaussians identified in the first step, it computes their splatting attributes of shape, color and opacity based on the camera's position. Third, the Gaussians are sorted by depth and their intersections with the rendering tiles are assessed, a process known as "key-duplication." The final output of this pre-processing pipeline is a sorted list of draw commands, organized per tile.

The conventional pre-processing stage performs a thorough computation of rasterization information for every single frame. Alternatively, \model introduces a Speculative Multi-frame Pre-processing method. This approach generates rasterization information for a batch of sequential frames by leveraging motion prediction and fine-grained Gaussian intersection judgment. Furthermore, a motion-adaptive scheduling mechanism is incorporated to enhance robustness in extreme scenes and guarantee rendering quality.

%In existing rendering pipelines, the pre-processing stage is performed thoroughly for every frame. 
%However, as we observe, during high-frame-rate rendering of a continuous scene, this process contains significant redundancy and leads to performance waste. Firstly, frustum culling requires calculating the entire model to determine visibility. In continuous rendering, however, the set of visible Gaussians remains largely consistent across adjacent frames due to the gradual viewpoint movement.
%\shengzhong{This paragraph should be a summary of what we propose in this subsection. Do not stop at the issues in existing pipelines. I suggest we compress the discussion on existing limitations, and further point out the summary of this part.}

\subsubsection{Motion Prediction}
To extend the validity of pre-processing results across multiple adjacent frames, we begin by analyzing how frame contents change across frames, as this is determined by the input viewpoint's movement.
We first establish a model to quantify the effect of viewpoint movement. We define three types of viewpoint motions that can occur in any instantaneous movement: \textit{panning}, \textit{zooming}, and \textit{rotation}. 
Let the world coordinate be denoted by the translation vector $T = (T_x, T_y, T_z)$ and the orientation by a $3 \times 3$ rotation matrix $R = (R_1, R_2, R_3)$, and inner parameter matrix $K$.
%To extend the validity of preprocessing results across multiple adjacent frames, we propose a "trace intersection judgment" method. Unlike conventional approaches that generate rasterization data solely for the current viewpoint, our method produces a unified set of rasterization information for a batch of subsequent frames. This set comprehensively includes all points required for any individual frame within the batch, plus a small number of redundant points to accommodate the needs of other frames. 

By measuring the changes in the translation vector T and rotation vector R of the input camera over a short time interval, we can estimate the camera motion across a few frames. Within this short time interval, the motion can be considered nearly constant and unique.
Suppose within the target time interval, the camera moves from position $T$ to $T'$s and $R$ to $R'$. The affine transformation that maps a Gaussian 3D point at $(x_g, y_g, z_g)$ to the pixel coordinates $(u, v)$ to $(u',v')$ can be derived from Equation~\ref{eq1}. We can then derive a motion vector $(\Delta u,\Delta v) = (u',v') - (u, v)$ with respect to the coordinate change.%\bo{you can save space for this equation by replacing those vectors with symbols and explain symbols inline.}
\begin{equation}
\label{eq1}
    \begin{bmatrix}u\\v\\1\end{bmatrix} = \frac{1}{z_c}K(R\cdot\begin{bmatrix}x_g\\y_g\\z_g\end{bmatrix} + T).
\end{equation}
%\shengzhong{The equation is not defining the $(\Delta u,\Delta v)$.}
This motion vector $(\Delta u,\Delta v)$ represents the pixel-level trajectory of a Gaussian center across a sequence of frames. The region swept by the Gaussian along this trajectory is termed the Gaussian trace. Although the projected shape of the Gaussian changes slightly due to rotation, we treat it as constant since its variation is negligible.

%\begin{equation}
%\begin{bmatrix}u'\\v'\\1\end{bmatrix} = \frac{1}{z_c'}K(R'\cdot\begin{bmatrix}x_g\\y_g\\z_g\end{bmatrix} + T + \Delta T)
%\end{equation}

\subsubsection{Fine-grained Intersection Detection}
\begin{figure}
    \centering
    \includegraphics[width=\linewidth]{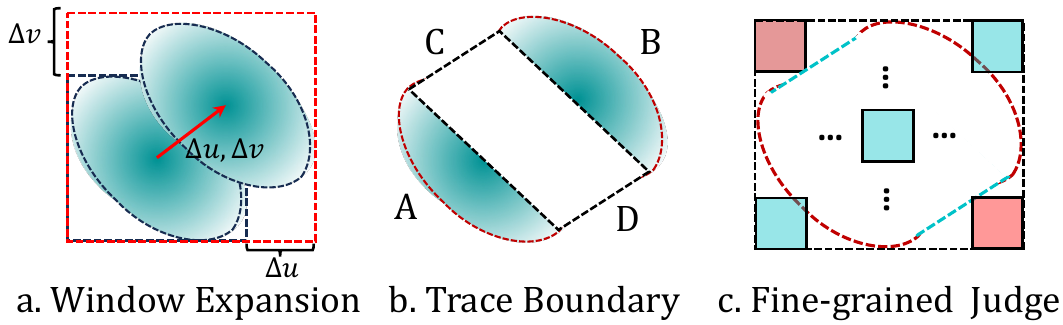}
    \caption{Steps to apply fine-grained intersection detection to a Gaussian trail.}
    \label{fig:intersec}
\end{figure}
Unlike previous methods that judge the intersection per Gaussian, \model determines tile intersection based on the entire swept area—the region encompassed by a Gaussian's motion along the vector $(\Delta u, \Delta v)$, as we call this the \textbf{Gaussian trail}. During rendering, each pixel iterates over all relevant Gaussians within the tile to compute its color. By predicting motion and judging intersection with the Gaussian trail, \model ensures that the rendered list includes all necessary Gaussians for a batch of frames.

To determine the intersection of a Gaussian trace, we first expand the bounding window by $\Delta u$ and $\Delta v$, as illustrated in Figure~\ref{fig:intersec}a. Subsequently, we perform intersection tests on all tiles overlapped by this expanded rectangle.
The boundary of a Gaussian trail comprises four segments: two half-ellipses centered at $(u, v)$ and $(u+\Delta u, v+\Delta v)$, labeled $A$ and $B$ in Figure~\ref{fig:intersec}b, and two parallel lines $C$ and $D$ that connect the tangent points of the two ellipses along the motion vector.
These four boundary segments can be mathematically estimated using 2D covariance $\Sigma$ derived from Gaussian features. The two half-ellipses are implicitly defined by the 2D inverse covariance matrix $\Sigma^{-1}$ of the Gaussian (as in Equation~\eqref{eq:sigma}) and a distribution clamp boundary parameter $p$. Specifically, each ellipse satisfies the Equation~\eqref{eq:ellipse}.
From this implicit function, we can also derive the equations for lines $C$ and $D$ (as in Equation~\eqref{eq:8}) by computing the tangent points $(x_t, y_t)$ and $(-x_t, -y_t)$ with respect to the motion vector $(\Delta u, \Delta v)$.   

\begin{equation}
\label{eq:sigma}
\Sigma^{-1} = \begin{pmatrix}
a & b \\
c & d
\end{pmatrix},
\end{equation}
\begin{equation}
\label{eq:ellipse}
    ax^2 + bxy + cy^2 = 2p ,
\end{equation}
\begin{equation}
\label{eq:8}
    \begin{cases} 
X(t) = x_t + t \cdot \Delta u \\ 
Y(t) = y_t + t \cdot \Delta v 
\end{cases}
, \quad t \in [0, 1].
\end{equation}

\noindent For each tile block in the expanded window, the intersection of two ellipses and lines is identified. The tile is valid for a Gaussian trace if: 1) The block intersects with any of the boundaries; 2) The block's center lies within the rectangle formed by the two ellipse centers.

\subsubsection{Motion Adaptive Adjustment}
%\bo{Scheduling is not clear. Maybe Motion Adaptive Window Adjustment?}
A fundamental prerequisite for \model is the similarity between adjacent frames. When motion becomes too drastic, the speculative preprocessing would incorporate excessive redundant information; therefore, a boundary condition is automatically set to avoid such extreme cases.
The displacement of a Gaussian point induced by camera motion generally depends on its projected depth, denoted as $z_c$ in Equation~\eqref{eq1}. The closer a Gaussian is to the camera, the more significantly it moves in response to camera motion. Thus, to avoid interference from Gaussians that are too close to the camera, we assess the validity of the motion speculation by only considering Gaussians whose depth is beyond a threshold $d$.

%\shengzhong{Why the rendering frame rate is 120fps? Besides, we should give the intuition on why the following is selected?}
We set the target rendering frame rate to \( F_{\text{target}} \), which is higher than the frame rates in our test cases, and initialize the speculation window to \( W_{\text{initial}} \) frames—meaning one key frame followed by \( W_{\text{initial}} \) subsequent sub-frames. If any Gaussian deeper than a threshold \( d \) exhibits movement \( M = \sqrt{\Delta u^2 + \Delta v^2} \) greater than the tile size \( l \), the motion is considered too drastic. In such cases, the window size is reduced to \( \left\lfloor W_{\text{initial}} \times \frac{l}{M} \right\rfloor \) to satisfy the constraint. Should the motion between two adjacent frames alone exceed the limit, resulting in \( \left\lfloor W_{\text{initial}} \times \frac{l}{M} \right\rfloor = 0 \), the speculation strategy is skipped for the current frame.
%\begin{equation}
%\begin{bmatrix}\Delta u\\\Delta v\end{bmatrix} \approx J \cdot \Delta T
%\end{equation}

%\begin{equation}
%\begin{bmatrix}\Delta u\\\Delta v\end{bmatrix} \approx \frac{1}{z_c} \begin{bmatrix}
%f_x & 0 & -\frac{f_x(x_c+\Delta x)}{z_c}\\
%0 & f_y & -\frac{f_y(y_c+\Delta y)}{z_c}
%\end{bmatrix} \cdot \begin{bmatrix}\Delta x\\\Delta y\\\Delta z\end{bmatrix}
%\end{equation}

\subsection{Inter-frame Caching}
\label{sec:3.2}
By leveraging a speculative pre-processing stage that generates rasterization information for a batch of frames, we can mitigate redundant computation. In vanilla 3DGS, the pre-processing stage contains three steps: 1) frustum culling, 2) feature computation, and 3) sorting and list generation. 
Only feature computation needs to be performed per frame, as the color, shape, and opacity of the Gaussians are highly viewpoint-dependent. For other steps, \model employs a caching mechanism to avoid redundant computations.

\subsubsection{Frustum Caching}
For a continuous sequence of frames, frustum culling only needs to be performed once, as the speculative pre-processing stage has already identified the required subset of Gaussians of the model. 
%\shengzhong{Subset of Gaussian or other things?}
Therefore, in subsequent frames, frustum culling can be omitted entirely. 
To implement this caching mechanism, \model calculates and stores the hashing indices of the visible Gaussians from the first frame in the batch. 
For the following frames, \model no longer needs to iterate through the entire set of Gaussians and compute their affine transformations to determine visibility. Instead, it directly utilizes the pre-culled set. This optimization yields greater performance benefits in large scenes, where the culled set typically represents only a small portion of the entire model.

\subsubsection{Sort Caching}
Regarding the sorting step, an intuitive fact is that the depth order of Gaussians remains mostly consistent across frames within a short time, as camera position and rotation change very little under high frame rates. This insight led to the design of our sort caching mechanism. 
In \model, the first frame in a batch performs sorting and list generation. After rasterization, this list is cached and reused for subsequent frames, as it already contains all the necessary rasterization information for the entire batch. Consequently, later frames skip the sorting stage and rasterize directly using the cached list.
%However, while skipping sorting saves time, reusing the list for multiple frames unavoidably introduces redundancy in the form of unnecessary draw calls for each frame. This redundancy has the most significant impact on the first frame. Since the first frame must execute the full rendering pipeline, the additional computational load from these redundant calls lowers the minimum frame rate and compromises the smoothness of the experience. %To allieviate this problem, \model modified the 
Since reusing the list for multiple frames unavoidably introduces redundancy, as each tile contains unneeded points from the current viewpoint, the fine-grained intersection and motion-adaptive scheduling control the redundancy to an acceptable level, as discussed in Section~\ref{sec:5_redun}. 
More importantly, sort caching also eliminates the need for repetitive tile intersection judgment and key-duplication in the preprocessing stage for subsequent frames, which significantly accelerates the overall rendering process.
\begin{figure}[htbp]
    \centering
    \includegraphics[width=1.04\linewidth]{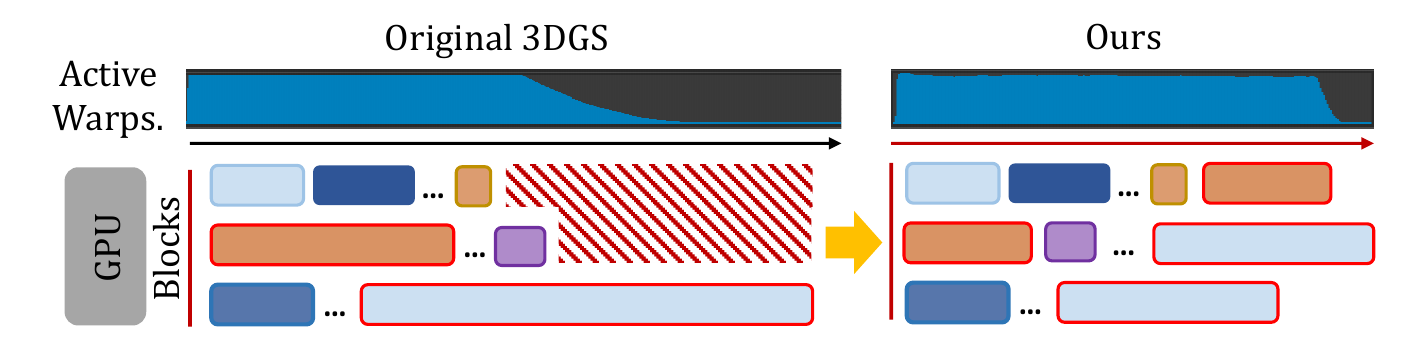}
    \caption{Task redistribution and efficient kernel improve rasterization performance. Active warps row shows an example result profiled from Nsight Compute.}
    \label{fig:split}
\end{figure}

\subsection{Load-Aware Task Splitting}
%\bo{Maybe load-aware? Load-balanced here may not be 100\% correct as our approach is somewhat heuristic.}
\label{sec:3.4}

In this section, we focus on performance bottleneck caused by tile load imbalance. As analyzed in Section~\ref{sec:motivate}, the distribution of the Gaussian model leads to an uneven computational workload of each rasterization tile in parallel rendering of the GPU, resulting in low utilization.

\noindent \textbf{Stalls in Parallel Rendering}  This issue stems from the fundamental execution strategy of the GPU. During rendering, each GPU unit—known as a Streaming Multiprocessor (SM)—concurrently launches multiple tile rasterization tasks (referred to as block launches). Each launched block is executed entirely on the same SM that launched it, meaning that the rendering task for a given tile is confined to a single SM. To fully utilize computational resources, all SMs on the GPU immediately launch a new block as soon as one is completed. This process can be simplified as shown in Figure~\ref{fig:split}, which illustrates 3 parallel compute units launching render blocks. However, since a small number of tasks often accounts for the majority of the computational overhead, this execution strategy can lead to stalls. Larger tasks in the distribution may become stuck on a single SM, delaying overall completion.

\noindent\textbf{Tile-level Splitting} To prevent large task blocks from stalling on a single Streaming Multiprocessor (SM), an intuitive approach is to split heavy tile tasks into smaller ones, allowing the GPU to schedule them across multiple SMs. In other words, for regions with high computational loads, we assign more SMs to collaboratively perform the computations, thereby accelerating the rendering process. However, this splitting cannot be done arbitrarily due to the order dependency in alpha blending. As described by Equation~\eqref{eq:3}, the process originally requires ordered computation, which cannot be parallelized at the pixel level.
To tackle such problem, we refactor the rasterization step in 3DGS. Specifically, to obtain color $C$ of a pixel by blending $N$ ordered points (splats) overlapping the pixel, we split the rendering task into $k$ subdivisions. These correspond to subsets of the total list of N points, denoted as $N_1,N_2...N_k$, The alpha blending process can then be performed in a batched manner, which can be formulated as follows:
\begin{align}
C &= \sum_{i=1}^{k} (\sum_{j=1}^{|N_i|} c_j \alpha_j T_j)  R_i, \\
R_i &= \prod_{m=1}^{i-1} (1 - A_m),A_m = \sum_{j=1}^{|N_m|} \alpha_j T_j.
\end{align}
In this formulation, each tile block now processes one of the $k$ subdivisions of size $N/k$ instead of handling the entire original task. Therefore, for tiles with a dense Gaussian distribution, our model splits the original rendering task into smaller subtasks that can be scheduled efficiently by the GPU. These subtasks can be allocated to multiple SMs for parallel rendering, thereby preventing the workload from stalling on a single SM and improving overall utilization. 

\noindent\textbf{Split Render Kernel} To implement in rasterization kernel, we obtain a load-aware splitting strategy. For a render list generated in the previous stage, the total computational load can be determined by its length $L$. The average load per tile is then $l = L / t$, where $t$ is the number of tiles. We identify any tile containing more than $2l$ Gaussians to rasterize as a large tile, and subsequently split it into smaller tiles targeting a size of $l$. %This process id done in tile identifying step, 
During rasterization, \model launches thread blocks equivalent to the total number of split tasks. Each task is assigned to a tile index $t$ and a slice index $k$. A thread block computes a portion of the expression  $\sum_{j=1}^{|N_k|} c_j \alpha_j T_j$  which corresponds to the part of the original task it is responsible for. After completing this partial computation, each block stores its slice color $C_k$ and slice radiance residual $A_k$ for the final image composition.

\section{Implementation}\label{sec:implementation}

We implemented the \model framework with $\sim$3K CUDA and Python code. The render pipeline is built via Pytorch~\cite{paszke2019pytorchimperativestylehighperformance} framework, and the rasterization kernel is compiled via Pytorch CUDA extension. To collect a sequential trace of rendering, we use the SIBR\_Viewers renderer to collect user interactive trace while viewing the Gaussian models. The block size of rendering is set to $16\times16$. The target frame rate $F_{\text{target}}$ is set to 120, threshold $d$ to $0.4$ and initial window size $W_{\text{initial}}$ to 4 frames. 

%\bo{Need evaluation results on the number hardcoded in the design, such as the ratio of key to sub frames, the number we divide each tile, and the fraction of tile we divide.}
\section{Evaluation}\label{sec:experiment}

\begin{table*}[htbp]
\centering
\caption{Average FPS across all traces on each model mentioned in Section~\ref{subsec:experiment-1}. The best result is marked as grey cell color. The improvement against the previous SOTA result is marked by $\uparrow$ after our result. Model size indicates the number of Gaussians in the model.}
\label{tab:fps}
\resizebox{\textwidth}{!}{%
\begin{tabular}{@{}lcccccccccc}
\toprule
& Truck & Train  & Counter & Garden & Playroom  & UAV-1 & UAV-2 & UAV-3 &UAV-4 &UAV-5\\%Scene-18&Scene-13&Scene-17&Scene-15&Scene-16\\
\cmidrule(lr){2-11}
Model Size& 2,541,226 & 1,026,508   &1,222,956  & 4,232,564 &2,546,116   &6,895,821 & 7,189,662 & 5,926,371& 8,324,723 &7,396,621 \\
\midrule
3DGS & 63.4 & 89.4   & 122.3  & 92.6 & 68.1 & 25.2 & 23.2 & 54.3 & 47.2 & 36.1 \\
ADR-GS & 425.2 &482.1  &498.1   & 204.1 & 593.2 &116.8  & 98.3  & 214.2 & 132.1 & 111.1 \\
Flash-GS & 478.3 & 528.3  & 606.3 & 201.1 & 634.2 & 129.2 & 113.1 & 241.1 & 167.3 & 132.3 \\
\midrule
Ours-w/o & 494.4 & 550.2  & 626.1 & 242.4 & 654.8 & 155.3 & 133.2 & 249.6 & 174.2 & 150.1 \\
Ours & \cellcolor{gray!20} 736.5 ($\uparrow 54\%$) & \cellcolor{gray!20} 892.2($\uparrow 68\%$) & \cellcolor{gray!20} 907.5 ($\uparrow 49\%$) & \cellcolor{gray!20} 295.5 ($\uparrow 46\%$) & \cellcolor{gray!20} 922.3 ($\uparrow 45\%$) & \cellcolor{gray!20} 241.5 ($\uparrow 83\%$) &  \cellcolor{gray!20} 202.5($\uparrow 78\%$) &  \cellcolor{gray!20} 378.5($\uparrow 64\%$)& \cellcolor{gray!20} 272.1 ($\uparrow 62\%$) & \cellcolor{gray!20} 217.3 ($\uparrow 65\%$)   \\

\bottomrule
\end{tabular}%
}
\end{table*}

\begin{table*}[htbp]
\centering
\caption{Render quality on video-captured datasets. The $*$ after ADR-GS means the model is retrained due to the need for the method. The other baselines use the same model trained with 3DGS.}
\label{tab:quality}
\resizebox{0.85\textwidth}{!}{%
\begin{tabular}{@{}l*{15}{c}@{}}
\toprule
 & \multicolumn{2}{c}{Truck} & \multicolumn{2}{c}{Train} & \multicolumn{2}{c}{UAV-1} & \multicolumn{2}{c}{UAV-2} & \multicolumn{2}{c}{UAV-3} & \multicolumn{2}{c}{UAV-4} & \multicolumn{2}{c}{UAV-5}  \\
\cmidrule(lr){2-3} \cmidrule(lr){4-5} \cmidrule(lr){6-7} \cmidrule(lr){8-9} \cmidrule(lr){10-11}  \cmidrule(lr){12-13}  \cmidrule(lr){14-15}
 & PSNR & SSIM& PSNR & SSIM & PSNR & SSIM  & PSNR & SSIM& PSNR & SSIM & PSNR & SSIM & PSNR & SSIM   \\
\midrule
3DGS & 28.21 & 0.858  & 25.82 & 0.872 & 30.15 & 0.935 &29.89 & 0.889 & 29.53 & 0.909 & 30.37 & 0.943   & 28.34 & 0.872  \\
ADR-GS* & 26.50 & 0.824 & 23.92 & 0.768 & 29.12 & 0.883& 29.31& 0.903& 27.34 & 0.878& 28.69 & 0.872  & 27.26 & 0.831\\
FlashGS & 28.19 & 0.856  & 25.78 & 0.869 & 30.09 & 0.931& 29.86 & 0.886   & 29.50 & 0.905 & 30.32 & 0.940 & 28.30 & 0.869  \\
Ours-Key & \cellcolor{gray!20}28.19 &\cellcolor{gray!20} 0.856 &\cellcolor{gray!20}25.78 & \cellcolor{gray!20}0.869 & \cellcolor{gray!20}30.09 &\cellcolor{gray!20} 0.931&\cellcolor{gray!20}29.86 &\cellcolor{gray!20}0.886 &\cellcolor{gray!20}29.50 &\cellcolor{gray!20}0.905 &\cellcolor{gray!20}30.32 &\cellcolor{gray!20}0.940  & \cellcolor{gray!20}28.30 & \cellcolor{gray!20}0.869   \\
Ours-Sub & \cellcolor{gray!20}28.16 &\cellcolor{gray!20} 0.852  &\cellcolor{gray!20}25.74 &\cellcolor{gray!20}0.867 &\cellcolor{gray!20}30.05 &\cellcolor{gray!20}0.926 &\cellcolor{gray!20}29.83&\cellcolor{gray!20}0.884 &\cellcolor{gray!20}29.46&\cellcolor{gray!20}0.903 &\cellcolor{gray!20}30.30&\cellcolor{gray!20}0.939  &\cellcolor{gray!20}28.29 &\cellcolor{gray!20}0.868 \\
\bottomrule   
\end{tabular}%
}
\end{table*}

\subsection{Experiment Setup}\label{subsec:experiment-1}
\noindent\textbf{Test Platform} We tested our method on the platform of a Nvidia Geforce RTX 5090 GPU and an AMD Ryzen 9 9950X CPU with Ubuntu 24.04 OS. 

\noindent\textbf{Datasets} To comprehensively evaluate our rendering method, we employ both the standard public 3DGS datasets and the self-collected UAV city reconstruction dataset, which features a larger real-world scene scale. This dataset will be open-sourced in the future. The standard 3DGS datasets include Tanks \& Temples~\cite{Knapitsch2017} dataset, MipNeRF-360~\cite{barron2022mip} dataset, and Deep Blending~\cite{hedman2018deep} dataset. MipNeRF-360 dataset includes four indoor and five outdoor scenes, each featuring a complex central object or area with a detailed background. The Tanks \& Temples and Deep Blending datasets include 4 extra scenes.

To assess sequential rendering performance, we collected a series of user traces consisting of camera parameters for each model. Specifically, we gathered 10 trajectories per model, each spanning 2000 consecutive frames. These traces were recorded via interaction with the Gaussian models from the aforementioned datasets using the SIBR\_Viewer. The camera movements encompass a full range of motion types—including panning, zooming, and rotation—at varying speeds. All traces were captured at 120 FPS, and the resolution is set to $1920\times1080$.

\noindent\textbf{Evaluation Metrics}
As a render method, we evaluate \model on both render efficiency and quality. We use the average frame rate (\ie FPS) as the efficiency metric.
To evaluate render quality, we employ two standard image quality metrics: Peak Signal-to-Noise Ratio (PSNR), Structural Similarity Index Measure (SSIM).

\subsubsection{Baselines} We compare \model with the baselines below:
\begin{itemize}
    \item \textbf{3DGS~\cite{kerbl20233d}:} We compare to the original 3DGS render method from the released source code of the 3D Gaussian splatting work.
    \item \textbf{ADR-GS~\cite{wang2024adr}:} We compare to the previous ADR-GS baseline, as we re-trained the model with released source code.
    \item \textbf{Flash-GS~\cite{feng2025flashgs}:} We compare to previous SOTA method in software 3DGS rendering optimization. The baseline is also implemented with the released source code.
\end{itemize}
% \red{To be added.}

\subsection{Render Efficiency Improvement}
\label{sec:framerate}
To evaluate the render speed, we compare our model against our baselines in average FPS across the viewpoint trace. The result is shown in Table~\ref{tab:fps}. \model reaches the highest average fps due to the redundancy cut down and balanced task. All results of \model across large scenes in UAV dataset reach over 200Fps, while other baselines fail to ensure 120Fps.
Notably, \model-w/o represents our model without the inter-frame caching strategy, which also represents the worst case for \model. It can be seen that \model-w/o still reaches over 120FPS across large scale scenes due to efficiency improvement by tile task scheduling. Compared to the vanilla 3DGS, \model reaches up to $10\times$ speed up. From scene prospective, for the shown scenes from the standard datasets, which generally have Gaussian numbers between $1-4$ million, the improvement of \model against previous SOTA baseline is around $50\%$. For larger scenes in the UAV dataset, which have Gaussian number over $7$ million, the improvement is generally over $60\%$ and can go up to over $80\%$. \model generally exhibits a larger advantage as Gaussian model scales up, since a more complex model also involves heavier pre-processing and sorting stages.
%\shengzhong{We mentioned GPU utilization quite a few times. Do we have results on the average GPU utilization achieved by each method? Besides, not many interesting observations and analysises are given here.}

\subsection{Render Quality}
To evaluate the rendering quality of our method, we built our testing pipeline upon the original 3DGS evaluation framework. Specifically, we captured the original camera trajectory for the source views of each model, aligning each rendered frame with its corresponding ground truth image. 
We utilize datasets extracted from videos, as it offers pose-continuous ground truth frames. For "truck" and "train" from Tanks \& Temples dataset, the original ground truth was recorded at 30 FPS, we interpolated the frame rate to 120 FPS. The same 120 FPS interpolation was applied to UAV dataset as well.

To validate the speculative pre-processing stage, we assessed the rendering quality of both key-frames and sub-frames independently. This was done by aligning the timing of key-frames and sub-frames with the timestamps of available ground truth images, enabling a frame-accurate quality evaluation. The results are summarized in Table~\ref{tab:quality}. ADR-GS's render quality degrades to original 3DGS due to its retrained model's lower accuracy. Flash-GS maintains comparable quality with only minimal metric degradation, which is attributable to its opacity-aware Gaussian culling. The key-frame rendering of \model adopts the same strategy, resulting in the same render quality to Flash-GS. For \model’s sub-frames, a slightly larger degradation occurs due to potential mispositioning of several Gaussians in the speculative process. Overall, the rendering quality degradation of \model is negligible compared to previous baseline models, demonstrating the effectiveness of the speculative design.

\subsection{Ablation Study}
\noindent\textbf{Technique Ablation} We evaluate the impact of the two major mechanisms in our method: the inter-frame caching and the load-aware task splitting. For a detailed analysis, we select one model from each standard dataset and three models from the UAV dataset. The results are presented in Table~\ref{tab:abla}. The variant labeled "Ours-w/cache" denotes our system without inter-frame caching. As the results show, this mechanism improves overall performance by up to 80\%. Similarly, "Ours-w/split" denotes the method without load-aware task splitting. The results indicate that this optimization to the rasterization stage contributes an overall performance improvement of over $10\%$.

\begin{table}[h]
\centering
\caption{Impact of major mechanism in \model.}
\label{tab:abla}
\resizebox{0.45\textwidth}{!}{%
\begin{tabular}{@{}lcccccc@{}}
\toprule
AvgFPS. &Garden & Truck & Playroom & UAV-1 & UAV-2 & UAV-3 \\
\midrule
Ours-Full & 295.5 & 736.5 & 922.3 & 241.5 & 202.5 & 378.5\\
Ours-w/cache $\downarrow$& 242.4&  494.4 & 654.8 & 155.3 &133.2 &249.6 \\
Ours-w/split$\downarrow$& 279.4 & 672.3 &840.6  &210.3 & 178.4 & 337.6\\
\bottomrule
\end{tabular}%
}
\end{table}

\noindent\textbf{Stage Acceleration Analysis} We further break down the rendering time to better understand the improvements. The modification made by \model to the rendering pipeline involves four parts in 3D Gaussian splatting (3DGS), namely pre-processing, sorting, tile identification, and rasterization. We compare the average acceleration of each stage against the original 3DGS implementation.
The results are shown in Table~\ref{tab:accele}. In the pre-processing stage, Flash-GS improves upon the original 3DGS by fusing feature computation and key duplication into a single kernel. The key-frame pre-processing in \model is slightly slower than Flash-GS due to the additional computation involved in speculative pre-processing. In contrast, the sub-frame processing in \model only requires computing the features of the specific Gaussian group selected during frustum culling in the key-frame, achieving a speed-up of $6.8\times$ compared to the original 3DGS.
As for the sorting stage, all methods utilize the \emph{RadixSort} method. The main improvement comes from the shorter rasterization list resulting from more fine-grained intersection checks. The key-frame processing in \model is slightly slower due to a longer list that includes extra Gaussians for subsequent frames. However, since sub-frame processing does not require sorting, the average speed-up for this stage reaches up to $7.2\times$ when averaged across \model.
In the rasterization stage, though \model computes slightly more redundant points, it still achieves up to $30\%$ acceleration compared to the previous SOTA baseline, and an average $5\times$ speed up compared to original 3DGS.
\begin{table}[h]
\centering
\caption{Stage acceleration compared to original 3DGS. }
\label{tab:accele}
\resizebox{0.95\linewidth}{!}{%
\begin{tabular}{@{}lcccc}
\toprule
Speedup $\uparrow$ & Pre-processing & Sorting & Tile Identify & Rasterization \\
\midrule
3DGS & $1.0\times$ & $1.0\times$ & $1.0\times$ & $1.0\times$ \\
FlashGS&  $3.2\times$ & $3.3\times$ & $1.7\times$ & $4.2\times$  \\
Ours-key& $3.1\times$ & $2.9\times$ & $1.3\times$   & \cellcolor{gray!20}$5.3\times$    \\
Ours-sub& \cellcolor{gray!20}$6.8\times$ & \cellcolor{gray!20}0* & \cellcolor{gray!20}0* & \cellcolor{gray!20}$5.3\times$ \\

\bottomrule
\end{tabular}%
}
\end{table}

\noindent\textbf{Redundancy Analysis}
\label{sec:5_redun}
To further evaluate the rendering performance, we analyze the render list generated during pre-processing to assess the impact of the additional computation in our speculative multi-frame pre-processing. As shown in Table~\ref{tab:list}, the average render list length, which represents the number of Gaussian-tile intersections, is significantly reduced in Flash-GS. This reduction is achieved through more precise intersection tests with ellipses, which substantially decreases the computational load. In contrast, \model's key-frame speculative pre-processing accounts for Gaussian motion across sub-frames, inevitably introducing some redundancy into the render list. However, through fine-grained intersection checks and motion-adaptive scheduling, our design successfully limits this redundancy to an acceptable level, which only increases by around $10\%$.

\begin{table}[h]
\centering
\caption{Render list length across models.}
\label{tab:list}
\resizebox{0.45\textwidth}{!}{%
\begin{tabular}{@{}lcccccc}
\toprule
List Len.($\times 10^3$) &Garden & Truck & Playroom & UAV-1 & UAV-2 & UAV-3 \\
\midrule
3DGS & 18,593 & 18,969 & 22,550 & 37,821 & 52,336 & 34,532\\
FlashGS& 6,682   &  4,124 & 3,818 &10,766 &12,101 & 8,416 \\
Ours-key& \cellcolor{gray!20}7,032  & \cellcolor{gray!20}4,289 & \cellcolor{gray!20}4,182 & \cellcolor{gray!20}11,222& \cellcolor{gray!20}12,886 & \cellcolor{gray!20}8,628\\
Ours-sub&  \cellcolor{gray!20}7,032 & \cellcolor{gray!20}4,289 & \cellcolor{gray!20}4,182 & \cellcolor{gray!20}11,222& \cellcolor{gray!20}12,886 & \cellcolor{gray!20}8,628\\

\bottomrule
\end{tabular}%
}
\end{table}
\section{Related Works}\label{sec:related}
%\bo{Need to mention existing works that leverage frame-redundancy either for 3dGS or NeRF.}
3DGS~\cite{kerbl20233d} has marked a significant advance in 3D reconstruction and neural renderin, and has become commonly used in interactive applications~\cite{qian20243dgs,shen2025gaussianshopvr,han2025ggs}.
Subsequently, model compression techniques~\cite{liu2024compgs,hanson2025speedy,ali2025compression,zhang2024lp,lee2024safeguardgs,hanson2025pup,sun2024f} have been proposed to optimize 3DGS models by reducing their size and the number of Gaussians. Other research efforts focus on enhancing the rendering efficiency of 3DGS. Some approaches address this from a software perspective~\cite{feng2025flashgs,ye2025gsplat,wang2024adr,kheradmand2025stochasticsplats,niemeyer2025radsplat,hou2024sort}, while others target hardware optimizations~\cite{feng2025lumina,li2025gaurast}. Several studies further improve 3D reconstruction and rendering by exploiting frame redundancy~\cite{feng2025lumina,chen2025you,li2025streamgs}. Notably, Lumina~\cite{feng2025lumina} is a hardware-oriented optimization work for 3DGS that specifically addresses redundancy in the sorting stage during rendering.
\section{Conclusion}\label{sec:conclusion}
This paper presents \model, a novel rendering pipeline that significantly accelerates 3D Gaussian Splatting for large-scale scenes. By introducing a speculative multi-frame pre-processing mechanism with inter-frame caching, \model effectively eliminates redundant computations across consecutive frames. It further addresses tile-level load imbalance through a dynamic task-splitting strategy and an optimized CUDA kernel. Extensive experiments demonstrate that \model achieves a speedup of up to 10× over vanilla 3DGS and up to 70\% over prior SOTA methods. %\bo{the remaining part can be removed to save space}, while maintaining high rendering fidelity. Future work could explore a more fine-grained and adaptive caching strategy for further improvement.

\section*{Acknowledgement}
This work was supported in part by National Key R\&D Program of China (No. 2023YFB4502400), in part by China NSF grant No. 2472278, 62441236, 62372296, 62432007, 2332014, 2332013, and U25A6024, in part by Fundamental and Interdisciplinary Disciplines Breakthrough Plan of the Ministry of Education of China (No. JYB2025XDXM103), in part by Alibaba Group through Alibaba Innovation Research Program, in part by Tencent Rhino Bird Key Research Project, and in part by Shanghai QiYuan Innovation Foundation.

{
   \small
   \bibliographystyle{ieeenat_fullname}
    \bibliography{main}

@String(TOG= {ACM Trans. Graph.})

@String(AAAI = {AAAI})

@String(VR   = {Vis. Res.})

@String(TOG   = {ACM TOG})

@article{kerbl20233d,
  title={3D Gaussian splatting for real-time radiance field rendering.},
  author={Kerbl, Bernhard and Kopanas, Georgios and Leimk{\"u}hler, Thomas and Drettakis, George},
  journal={ACM Trans. Graph.},
  volume={42},
  number={4},
  pages={139--1},
  year={2023}
}

@article{mildenhall2021nerf,
  title={Nerf: Representing scenes as neural radiance fields for view synthesis},
  author={Mildenhall, Ben and Srinivasan, Pratul P and Tancik, Matthew and Barron, Jonathan T and Ramamoorthi, Ravi and Ng, Ren},
  journal={Communications of the ACM},
  volume={65},
  number={1},
  pages={99--106},
  year={2021},
  publisher={ACM New York, NY, USA}
}

@inproceedings{liu2024compgs,
  title={Compgs: Efficient 3d scene representation via compressed gaussian splatting},
  author={Liu, Xiangrui and Wu, Xinju and Zhang, Pingping and Wang, Shiqi and Li, Zhu and Kwong, Sam},
  booktitle={Proceedings of the 32nd ACM International Conference on Multimedia},
  pages={2936--2944},
  year={2024}
}

@article{ali2025compression,
  title={Compression in 3d gaussian splatting: A survey of methods, trends, and future directions},
  author={Ali, Muhammad Salman and Zhang, Chaoning and Cagnazzo, Marco and Valenzise, Giuseppe and Tartaglione, Enzo and Bae, Sung-Ho},
  journal={arXiv preprint arXiv:2502.19457},
  year={2025}
}

@inproceedings{hanson2025speedy,
  title={Speedy-splat: Fast 3d gaussian splatting with sparse pixels and sparse primitives},
  author={Hanson, Alex and Tu, Allen and Lin, Geng and Singla, Vasu and Zwicker, Matthias and Goldstein, Tom},
  booktitle={Proceedings of the Computer Vision and Pattern Recognition Conference},
  pages={21537--21546},
  year={2025}
}

@article{zhang2024lp,
  title={Lp-3dgs: Learning to prune 3d gaussian splatting},
  author={Zhang, Zhaoliang and Song, Tianchen and Lee, Yongjae and Yang, Li and Peng, Cheng and Chellappa, Rama and Fan, Deliang},
  journal={Advances in Neural Information Processing Systems},
  volume={37},
  pages={122434--122457},
  year={2024}
}

@inproceedings{wang2024adr,
  title={Adr-gaussian: Accelerating gaussian splatting with adaptive radius},
  author={Wang, Xinzhe and Yi, Ran and Ma, Lizhuang},
  booktitle={SIGGRAPH Asia 2024 Conference Papers},
  pages={1--10},
  year={2024}
}

@inproceedings{feng2025flashgs,
  title={Flashgs: Efficient 3d gaussian splatting for large-scale and high-resolution rendering},
  author={Feng, Guofeng and Chen, Siyan and Fu, Rong and Liao, Zimu and Wang, Yi and Liu, Tao and Hu, Boni and Xu, Linning and Pei, Zhilin and Li, Hengjie and others},
  booktitle={Proceedings of the Computer Vision and Pattern Recognition Conference},
  pages={26652--26662},
  year={2025}
}

@article{ye2025gsplat,
  title={gsplat: An open-source library for Gaussian splatting},
  author={Ye, Vickie and Li, Ruilong and Kerr, Justin and Turkulainen, Matias and Yi, Brent and Pan, Zhuoyang and Seiskari, Otto and Ye, Jianbo and Hu, Jeffrey and Tancik, Matthew and others},
  journal={Journal of Machine Learning Research},
  volume={26},
  number={34},
  pages={1--17},
  year={2025}
}

@inproceedings{chen2025you,
  title={You Only Render Once: Enhancing Energy and Computation Efficiency of Mobile Virtual Reality},
  author={Chen, Xingyu and Fang, Xinmin and Zhang, Shuting and Zhang, Xinyu and He, Liang and Li, Zhengxiong},
  booktitle={Proceedings of the 23rd Annual International Conference on Mobile Systems, Applications and Services},
  pages={263--276},
  year={2025}
}

@article{kheradmand2025stochasticsplats,
  title={StochasticSplats: Stochastic Rasterization for Sorting-Free 3D Gaussian Splatting},
  author={Kheradmand, Shakiba and Vicini, Delio and Kopanas, George and Lagun, Dmitry and Yi, Kwang Moo and Matthews, Mark and Tagliasacchi, Andrea},
  journal={arXiv preprint arXiv:2503.24366},
  year={2025}
}

@article{li2025gaurast,
  title={GauRast: Enhancing GPU Triangle Rasterizers to Accelerate 3D Gaussian Splatting},
  author={Li, Sixu and Keller, Ben and Lin, Yingyan Celine and Khailany, Brucek},
  journal={arXiv preprint arXiv:2503.16681},
  year={2025}
}

@article{li2025streamgs,
  title={StreamGS: Online Generalizable Gaussian Splatting Reconstruction for Unposed Image Streams},
  author={Li, Yang and Wang, Jinglu and Chu, Lei and Li, Xiao and Kao, Shiu-hong and Chen, Ying-Cong and Lu, Yan},
  journal={arXiv preprint arXiv:2503.06235},
  year={2025}
}

@inproceedings{feng2025lumina,
  title={Lumina: Real-Time Neural Rendering by Exploiting Computational Redundancy},
  author={Feng, Yu and Lin, Weikai and Cheng, Yuge and Liu, Zihan and Leng, Jingwen and Guo, Minyi and Chen, Chen and Sun, Shixuan and Zhu, Yuhao},
  booktitle={Proceedings of the 52nd Annual International Symposium on Computer Architecture},
  pages={1925--1939},
  year={2025}
}

@inproceedings{niemeyer2025radsplat,
  title={Radsplat: Radiance field-informed gaussian splatting for robust real-time rendering with 900+ fps},
  author={Niemeyer, Michael and Manhardt, Fabian and Rakotosaona, Marie-Julie and Oechsle, Michael and Duckworth, Daniel and Gosula, Rama and Tateno, Keisuke and Bates, John and Kaeser, Dominik and Tombari, Federico},
  booktitle={2025 International Conference on 3D Vision (3DV)},
  pages={134--144},
  year={2025},
  organization={IEEE}
}

@article{liu2025omnireason,
  title={OmniReason: A Temporal-Guided Vision-Language-Action Framework for Autonomous Driving},
  author={Liu, Pei and Ning, Qingtian and Lu, Xinyan and Liu, Haipeng and Ma, Weiliang and She, Dangen and Jia, Peng and Lang, Xianpeng and Ma, Jun},
  journal={arXiv preprint arXiv:2509.00789},
  year={2025}
}

@inproceedings{mallick2024taming,
  title={Taming 3dgs: High-quality radiance fields with limited resources},
  author={Mallick, Saswat Subhajyoti and Goel, Rahul and Kerbl, Bernhard and Steinberger, Markus and Carrasco, Francisco Vicente and De La Torre, Fernando},
  booktitle={SIGGRAPH Asia 2024 Conference Papers},
  pages={1--11},
  year={2024}
}

@inproceedings{sun2025streaming,
  title={Streaming 3DGS Virtual Worlds in 6DoF over Next-Generation Networks},
  author={Sun, Yuan-Chun},
  booktitle={Proceedings of the 33rd ACM International Conference on Multimedia},
  pages={13561--13565},
  year={2025}
}

@inproceedings{hess2025splatad,
  title={Splatad: Real-time lidar and camera rendering with 3d gaussian splatting for autonomous driving},
  author={Hess, Georg and Lindstr{\"o}m, Carl and Fatemi, Maryam and Petersson, Christoffer and Svensson, Lennart},
  booktitle={Proceedings of the Computer Vision and Pattern Recognition Conference},
  pages={11982--11992},
  year={2025}
}

@inproceedings{zhou2024drivinggaussian,
  title={Drivinggaussian: Composite gaussian splatting for surrounding dynamic autonomous driving scenes},
  author={Zhou, Xiaoyu and Lin, Zhiwei and Shan, Xiaojun and Wang, Yongtao and Sun, Deqing and Yang, Ming-Hsuan},
  booktitle={Proceedings of the IEEE/CVF conference on computer vision and pattern recognition},
  pages={21634--21643},
  year={2024}
}

@article{kwon2025realistic,
  title={Realistic and Interactive Virtual Museum Representation Using 3D Gaussian Splatting},
  author={Kwon, Ohyang and Yu, Jeongmin},
  journal={ISPRS Annals of the Photogrammetry, Remote Sensing and Spatial Information Sciences},
  pages={185--192},
  year={2025},
  publisher={Copernicus Publications G{\"o}ttingen, Germany}
}

@inproceedings{barron2022mip,
  title={Mip-nerf 360: Unbounded anti-aliased neural radiance fields},
  author={Barron, Jonathan T and Mildenhall, Ben and Verbin, Dor and Srinivasan, Pratul P and Hedman, Peter},
  booktitle={Proceedings of the IEEE/CVF conference on computer vision and pattern recognition},
  pages={5470--5479},
  year={2022}
}

@inproceedings{shi20253d,
  title={3D Gaussian-based immersive media streaming in networked extended reality},
  author={Shi, Yuang},
  booktitle={Proceedings of the 16th ACM Multimedia Systems Conference},
  pages={356--360},
  year={2025}
}

@inproceedings{lu2024scaffold,
  title={Scaffold-gs: Structured 3d gaussians for view-adaptive rendering},
  author={Lu, Tao and Yu, Mulin and Xu, Linning and Xiangli, Yuanbo and Wang, Limin and Lin, Dahua and Dai, Bo},
  booktitle={Proceedings of the IEEE/CVF Conference on Computer Vision and Pattern Recognition},
  pages={20654--20664},
  year={2024}
}

@inproceedings{wu2025advancing,
  title={Advancing Immersive Content Delivery with Dynamic 3D Gaussian Splatting},
  author={Wu, Nan and Lin, Weikai and Cheng, Ruizhi and Chen, Bo and Zhu, Yuhao and Nahrstedt, Klara and Han, Bo},
  booktitle={Proceedings of the 26th International Workshop on Mobile Computing Systems and Applications},
  pages={109--114},
  year={2025}
}

@article{Knapitsch2017,
    author    = {Arno Knapitsch and Jaesik Park and Qian-Yi Zhou and Vladlen Koltun},
    title     = {Tanks and Temples: Benchmarking Large-Scale Scene Reconstruction},
    journal   = {ACM Transactions on Graphics},
    volume    = {36},
    number    = {4},
    year      = {2017},
}

@article{hedman2018deep,
  title={Deep blending for free-viewpoint image-based rendering},
  author={Hedman, Peter and Philip, Julien and Price, True and Frahm, Jan-Michael and Drettakis, George and Brostow, Gabriel},
  journal={ACM Transactions on Graphics (ToG)},
  volume={37},
  number={6},
  pages={1--15},
  year={2018},
  publisher={ACM New York, NY, USA}
}

@article{lee2024safeguardgs,
  title={SafeguardGS: 3D Gaussian Primitive Pruning While Avoiding Catastrophic Scene Destruction},
  author={Lee, Yongjae and Zhang, Zhaoliang and Fan, Deliang},
  journal={arXiv preprint arXiv:2405.17793},
  year={2024}
}

@inproceedings{hanson2025pup,
  title={Pup 3d-gs: Principled uncertainty pruning for 3d gaussian splatting},
  author={Hanson, Alex and Tu, Allen and Singla, Vasu and Jayawardhana, Mayuka and Zwicker, Matthias and Goldstein, Tom},
  booktitle={Proceedings of the Computer Vision and Pattern Recognition Conference},
  pages={5949--5958},
  year={2025}
}

@inproceedings{chen20253dgv,
  title={3DGV: 3D Gaussian Splatting-Based Holographic Video Streaming over Wireless Networks},
  author={Chen, Zihao and Zou, Longhao and Tao, Xiaofeng},
  booktitle={2025 IEEE International Symposium on Broadband Multimedia Systems and Broadcasting (BMSB)},
  pages={1--6},
  year={2025},
  organization={IEEE}
}

@misc{paszke2019pytorchimperativestylehighperformance,
      title={PyTorch: An Imperative Style, High-Performance Deep Learning Library}, 
      author={Adam Paszke and Sam Gross and Francisco Massa and Adam Lerer and James Bradbury and Gregory Chanan and Trevor Killeen and Zeming Lin and Natalia Gimelshein and Luca Antiga and Alban Desmaison and Andreas Köpf and Edward Yang and Zach DeVito and Martin Raison and Alykhan Tejani and Sasank Chilamkurthy and Benoit Steiner and Lu Fang and Junjie Bai and Soumith Chintala},
      year={2019},
      eprint={1912.01703},
      archivePrefix={arXiv},
      primaryClass={cs.LG},
      url={https://arxiv.org/abs/1912.01703}, 
}

@inproceedings{qian20243dgs,
  title={3dgs-avatar: Animatable avatars via deformable 3d gaussian splatting},
  author={Qian, Zhiyin and Wang, Shaofei and Mihajlovic, Marko and Geiger, Andreas and Tang, Siyu},
  booktitle={Proceedings of the IEEE/CVF conference on computer vision and pattern recognition},
  pages={5020--5030},
  year={2024}
}

@inproceedings{shen2025gaussianshopvr,
  title={GaussianShopVR: Facilitating Immersive 3D Authoring Using Gaussian Splatting in VR},
  author={Shen, Yulin and Li, Boyu and Huang, Jiayang and Yip, David and Wang, Zeyu},
  booktitle={Proceedings of the 38th Annual ACM Symposium on User Interface Software and Technology},
  pages={1--14},
  year={2025}
}

@inproceedings{han2025ggs,
  title={Ggs: Generalizable gaussian splatting for lane switching in autonomous driving},
  author={Han, Huasong and Zhou, Kaixuan and Long, Xiaoxiao and Wang, Yusen and Xiao, Chunxia},
  booktitle={Proceedings of the AAAI Conference on Artificial Intelligence},
  volume={39},
  number={3},
  pages={3329--3337},
  year={2025}
}

@inproceedings{sun2024f,
  title={F-3dgs: Factorized coordinates and representations for 3d gaussian splatting},
  author={Sun, Xiangyu and Lee, Joo Chan and Rho, Daniel and Ko, Jong Hwan and Ali, Usman and Park, Eunbyung},
  booktitle={Proceedings of the 32nd ACM International Conference on Multimedia},
  pages={7957--7965},
  year={2024}
}

@article{hou2024sort,
  title={Sort-free gaussian splatting via weighted sum rendering},
  author={Hou, Qiqi and Rauwendaal, Randall and Li, Zifeng and Le, Hoang and Farhadzadeh, Farzad and Porikli, Fatih and Bourd, Alexei and Said, Amir},
  journal={arXiv preprint arXiv:2410.18931},
  year={2024}
}
}
\newpage
\clearpage
\setcounter{page}{1}
\maketitlesupplementary

\begin{algorithm}[t]
\caption{Identify Tile Ranges}
\label{alg:1}
\KwData{Render length $L$, point list $point\_list\_keys$, ranges array $ranges$, task counter $task\_count$, tile number $tile\_number$}
\KwResult{Updated $ranges$ array, render label $label$.}
$idx \gets global\ thread\ index;$

$k \gets gaussian\ counts\ threshold;$

$key \gets point\_list\_keys[idx];$

$currtile \gets key \gg 32$\;
$prevtile \gets 0$\;

\If{$idx = 0$}{
    $ranges[currtile].x \gets 0;$
    
    $task\_count \gets tile\_number;$
}
\Else{
    $prevtile \gets point\_list\_keys[idx-1] \gg 32$\;
    
    \If{$currtile \neq prevtile$}{
        $ranges[prevtile].y \gets idx;$
        
        $ranges[currtile].x \gets idx;$
    }
}

\If{$idx = L-1$}{
    $ranges[currtile].y \gets L;$
}

Synchronize threads,\ split the task.\;

\If{$currtile \neq prevtile$ \textbf{and} $idx \neq 0$}{
    $start \gets ranges[currtile].x;$
    
    $end \gets ranges[currtile].y;$

    $task\_size \gets end-start;$
    
    \If{$end - start > k$}{
        $split\_num \gets \lfloor task\_size / k \rfloor;$
        
        $ranges[currtile].y \gets start + (task\_size-split\_num*k);$
        
        \For{$i \gets 1$ \textbf{to} $split\_num$}{
            $offset \gets \text{atomicAdd}(task\_count, 1);$
            
            Write additional task range to $ranges[offset]$.
            
            Write tile information $currtile$ and split information $i$ to $label[offset]$.
        }
    }
}
\end{algorithm}

%\begin{algorithm}[H]
%\caption{Rasterization}
%\label{alg:1}
%\KwData{point list $P$, feature $F$, ranges $R$, labels $L$.}
%\KwResult{image $I$, radiance map $T$}

%$idx \gets block\_index;$

%$task\_flag \gets idx\%tile\_number;$
%\end{algorithm}
\subsection{Detailed Implementation}
\subsubsection{Dataset Details}
The UAV City Dataset is designed for benchmarking city reconstruction tasks using UAV-captured data. The dataset is constructed by flying a UAV around urban areas and extracting frames from recorded videos. It comprises 16 large-scale \textbf{real-world} scenes, each covering one to two city blocks, with a typical area exceeding 20,000 m². For each scene, the processed 3DGS dataset includes approximately 800 images along with the original video clip recorded at 30 FPS.

Since real-world scenes contain more complex geometries and textures—such as trees and grasslands—this dataset offers a more realistic simulation of practical applications compared to previous graphically modeled 3D city datasets like UrbanScene3D. This dataset will be open-sourced in recent months.

\subsubsection{Task Splitting Implementation}
We provide a detailed explanation of our design in Section~\ref{sec:3.4} of the paper. The task splitting strategy is implemented by modifying the Tile-Identify step of 3DGS. As outlined in Algorithm~\ref{alg:1}, after generating the tile range for rasterization, we check whether the number of Gaussians in each tile exceeds a threshold $k$. If so, the tile task is split with respect to $k$, generating new subtasks. To identify the tile and the number of splits during rasterization, additional task information of tile index and partition number is stored in a $label$ array. The total task number is reformed to $task\_count$.

During the rasterization stage, the kernel launches $task\_count$ blocks. Blocks with an index below the original tile count proceed with rasterization as usual. For the additional blocks corresponding to split tasks, each one retrieves its respective parameters from the $label$ array.  Based on the tile index and split index provided by the $label$ array, the output is written to the corresponding pixel and slice of the image tensor and radiance map to store the values of $R$ and $A$ as defined in Equation~\eqref{eq:8} of the paper.

A redundancy introduced by our task-splitting kernel involves the early-termination mechanism of rasterization. In standard 3DGS, this mechanism halts alpha blending once the accumulated radiance becomes sufficiently small, which can be intuitively understood as the light being absorbed by preceding Gaussians along the path. Consequently, in our split rendering approach, the latter portions of a split task may become unnecessary. To mitigate this, we employ two key adjustments: first, the early-termination threshold is set higher for these additional task blocks; second, during task splitting, we allocate a larger size to the latter tasks, the first sub-task is set to a size of $task\_size-split\_num*k$ rather than $k$. Generally, this approach introduces some additional computational overhead compared to the original method. Nevertheless, the overall acceleration achieved still results in performance improvement.

\subsection{Speculative Pre-processing Details}
\begin{figure}
    \centering
    \includegraphics[width=\linewidth]{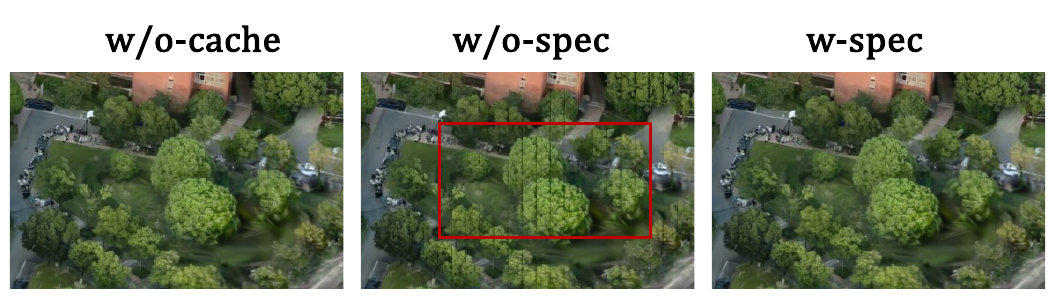}
    \caption{The speculative process includes the needed Gaussian in subsequent frames and avoids fault effects.}
    \label{fig:app4}
\end{figure}

We will further demonstrate the effectiveness and robustness of our speculative multi-frame pre-processing design in Section~\ref{sec:3.2}. Figure~\ref{fig:app4} shows an example of the last frame in a cached batch, which illustrates the impact of our approach. The first column under w/o-cache shows the result without inter-frame caching—reflecting the original 3DGS pipeline. The second column presents the result of applying inter-frame caching directly without speculative pre-processing. As shown, stride-like visual artifacts appear due to missing Gaussians in certain tiles caused by viewpoint movement. The third column displays the result of \model, where the speculative pre-processing stage successfully incorporates all necessary points across the frame batch, as the render result remains consistent with the original pipeline.

\begin{table}[h]
\centering
\caption{The impact of speculation window size.}
\label{tab:app2}
\resizebox{0.45\textwidth}{!}{%
\begin{tabular}{@{}lcccccc}
\toprule
UAV-1 &FlashGS& None & W-2 & W-4  & W-8 & W-16  \\
\midrule
Render Len.$\times10^3$&10,766 &10,766&10,982&11,222 &15,482 & 25,934  \\
Max Latency(ms) &7.7& 6.3& 6.4 & 6.6 & 9.1 &  14.1 \\
\toprule
UAV-2 &FlashGS& None & W-2 & W-4  & W-8 & W-16  \\
\midrule
Render Len.$\times10^3$&12,101 &12,101&12,382 &12,886 & 15,926 & 24,621  \\
Max Latency(ms) & 8.6& 7.1 & 7.2 & 7.5 & 10.9 & 13.4 \\
 
\bottomrule
\end{tabular}%
}
\end{table}

We also conducted an ablation study on the speculation window size to evaluate its impact on performance and robustness. As an example, Table~\ref{tab:app2} presents the corresponding results. Rows labeled W-2 to W-16 show the outcomes when the speculation window size is set from 2 to 16. It can be observed that the rendering length increases nonlinearly as the window size grows, which is attributed to the quadratic decrease in image tile overlap across frames. Although the rendering quality remains consistent even with a window size of 16, the maximum latency increases significantly if the window size is set too large, and this will impact the rendering smoothness. Therefore, we set the initial window size to 4 and designed a Motion Adaptive Adjustment mechanism to dynamically control the window size. A more fine-grained adaptive window strategy could be developed in the future as a potential improvement to our work.

\subsection{Compatibility With Pruning Method}
In this section, we discuss the adaptability of render pipeline optimization and model pruning. Although model pruning suffers from certain deficiencies that limit its practicality, as outlined in Section 1, we demonstrate that it can be effectively combined with our rendering optimization method to yield situational advantages. Even in these scenarios, our approach maintains superior performance over previous baselines. To evaluate this, we applied the Compact-3DGS method to retrain and prune models on the UAV dataset. As shown in Table~\ref{tab:app1}, UAV1-3 represents the original large-scale model, while UAV1-3P denotes the pruned versions. The results indicate that our method still retains an advantage over the baselines. Furthermore, the combination of model pruning and our method exhibits promising potential for achieving smooth rendering in larger-scale scenes.
\begin{table}[h]
\centering
\caption{Render speed across original models and pruned models.}
\label{tab:app1}
\resizebox{0.45\textwidth}{!}{%
\begin{tabular}{@{}lcccccc}
\toprule
AvgFPS.& UAV-1 & UAV-2 & UAV-3  &UAV-1P & UAV-2P& UAV-3P \\
\cmidrule{2-7}
Size$\times10^3$&6,895&7,189&5,926 & 2,671 &3,269 &2,241 \\
\midrule

3DGS &25.2&23.2 &54.3 & 49.2 & 44.6 & 73.2 \\
Flash-GS &129.2 &113.1 & 241.1& 296.6&278.2 & 436.9\\
Ours &\cellcolor{gray!20}241.5 & \cellcolor{gray!20}202.5& \cellcolor{gray!20}378.5 & \cellcolor{gray!20}489.4& \cellcolor{gray!20}448.3& \cellcolor{gray!20}746.2\\

\bottomrule
\end{tabular}%
}
\end{table}

\subsection{Visual Effects}
Finally, we present visual comparisons between our method and the baseline approaches. The figure below shows rendering examples from random viewpoints. It can be observed that our results are largely consistent with those of Flash‑GS, which is expected as our method also employs opacity‑aware culling in the pre‑processing stage. In contrast, when compared with 3DGS, our approach and FlashGS exhibit minor distortions in detailed image regions, often manifesting as color drift in uniform areas. We attribute these artifacts to the opacity‑aware culling strategy and consider this issue a potential direction for future optimization.

%\section{Visual Effects}
%We present example visual effects of \model and %baselines in the part below.
\begin{figure*}
    \centering
    \includegraphics[width=\linewidth]{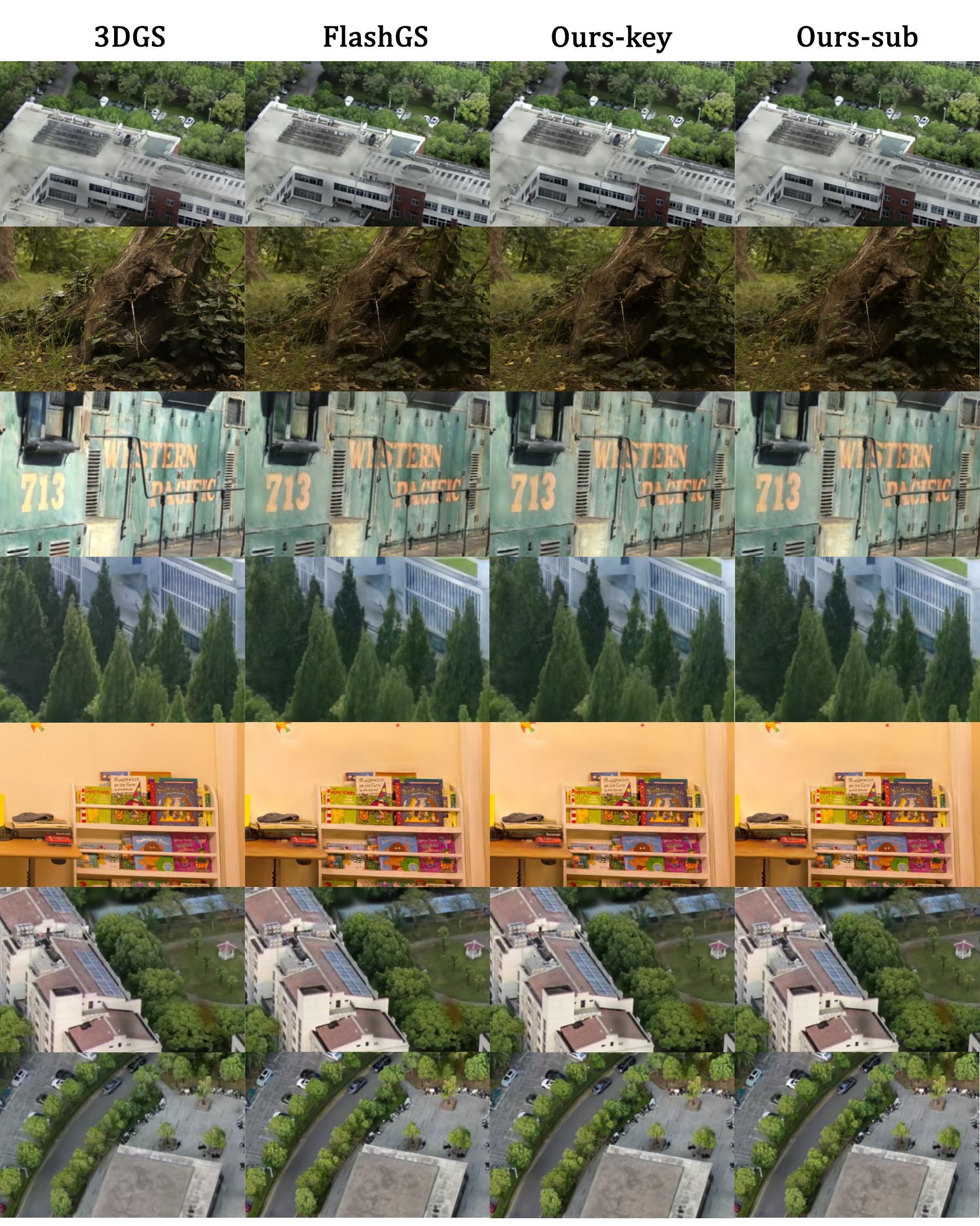}
    %\caption{Caption}
    \label{fig:placeholder}
\end{figure*}

\begin{figure*}
    \centering
    \includegraphics[width=\linewidth]{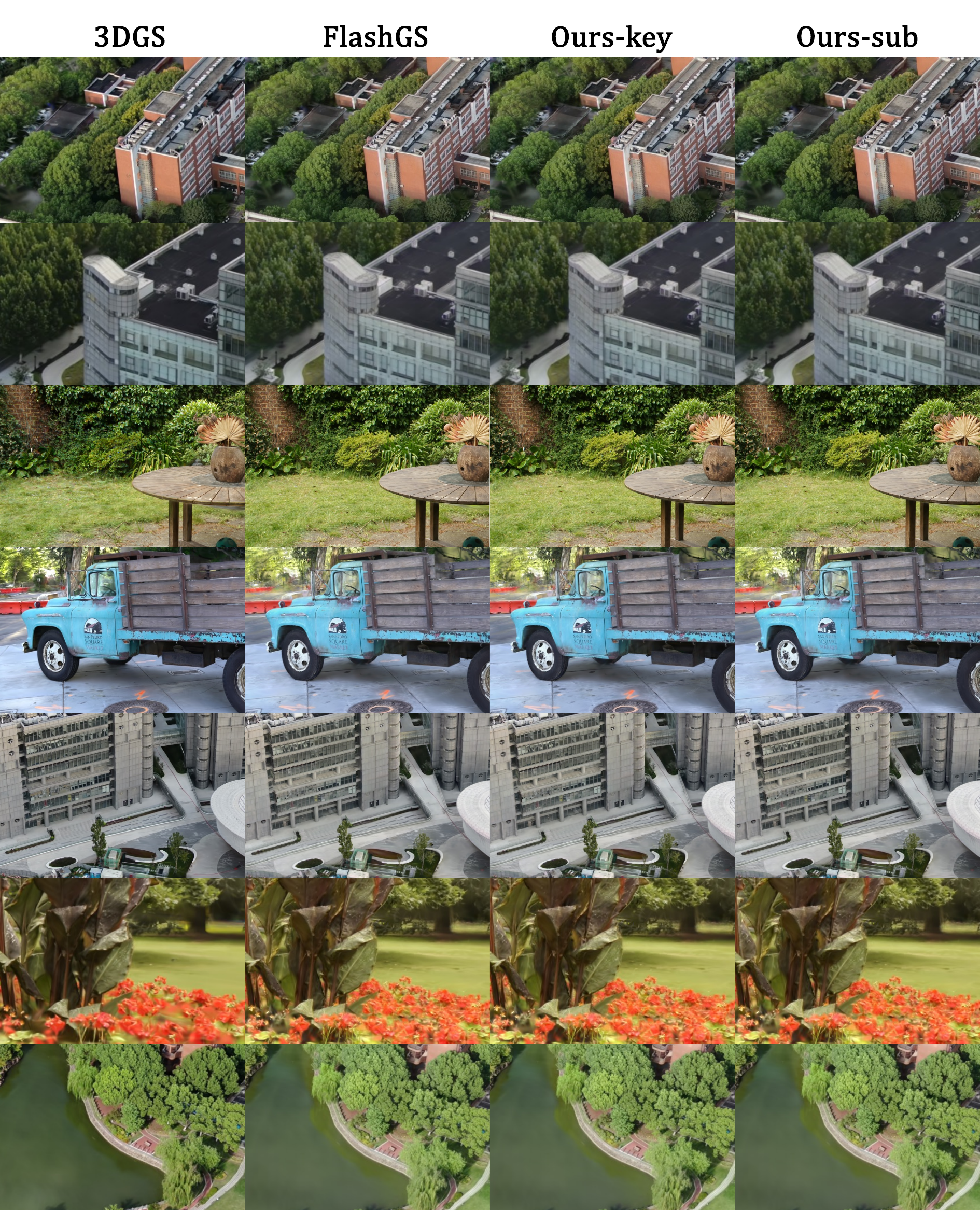}
    %\caption{Caption}
    \label{fig:placeholder}
\end{figure*}

\end{document}